\documentclass[letterpaper]{article} % DO NOT CHANGE THIS
\usepackage{aaai23}  % DO NOT CHANGE THIS
\usepackage{times}  % DO NOT CHANGE THIS
\usepackage{helvet}  % DO NOT CHANGE THIS
\usepackage{courier}  % DO NOT CHANGE THIS
\usepackage[hyphens]{url}  % DO NOT CHANGE THIS
\usepackage{graphicx} % DO NOT CHANGE THIS
\usepackage{natbib}  % DO NOT CHANGE THIS AND DO NOT ADD ANY OPTIONS TO IT
\usepackage{caption} % DO NOT CHANGE THIS AND DO NOT ADD ANY OPTIONS TO IT
\usepackage{algorithm}
\usepackage{algorithmic}

\usepackage{newfloat}
\usepackage{listings}
\DeclareCaptionStyle{ruled}{labelfont=normalfont,labelsep=colon,strut=off} % DO NOT CHANGE THIS
\floatstyle{ruled}
\newfloat{listing}{tb}{lst}{}
\floatname{listing}{Listing}
\title{Meta Learning in Decentralized Neural Networks: \\Towards More General AI}
\author {
    Yuwei Sun\textsuperscript{\rm 1,\rm 2}
}
\affiliations {
    \textsuperscript{\rm 1} The University of Tokyo\\
    \textsuperscript{\rm 2} RIKEN\\
    ywsun@g.ecc.u-tokyo.ac.jp
}

\usepackage{bibentry}
\begin{document}

\maketitle

\begin{abstract}
Meta-learning usually refers to a learning algorithm that learns from other learning algorithms. The problem of uncertainty in the predictions of neural networks shows that the world is only partially predictable and a learned neural network cannot generalize to its ever-changing surrounding environments. Therefore, the question is how a predictive model can represent multiple predictions simultaneously. We aim to provide a fundamental understanding of learning to learn in the contents of Decentralized Neural Networks (Decentralized NNs) and we believe this is one of the most important questions and prerequisites to building an autonomous intelligence machine. To this end, we shall demonstrate several pieces of evidence for tackling the problems above with Meta Learning in Decentralized NNs. In particular, we will present three different approaches to building such a decentralized learning system: (1) learning from many replica neural networks, (2) building the hierarchy of neural networks for different functions, and (3) leveraging different modality experts to learn cross-modal representations.
\end{abstract}

\section{Progress to Date} 

Common sense is not just facts but a collection of models of the world. The global workspace theory \cite{Baars1988} demonstrated that in the human brain, multiple neural network models cooperate and compete in solving problems via a shared feature space for common knowledge sharing, which is called the global workspace (GW). Within such a learning framework, using different kinds of metadata about individual neural networks such as measured performance and learned representations, shows the potential to learn, select, or combine different learning algorithms to efficiently solve a new task. The learned knowledge or representations from different neural network areas are leveraged for reasoning and planning. Therefore, we termed this research direction as Meta Learning in Decentralized Neural Networks which studies how a meta agent can solve novel tasks by observing and leveraging the world models built by these individual neural networks. We present the three different approaches to building such a decentralized learning system: (1) learning from many replica neural networks (2) building the hierarchy of neural networks, and (3) leveraging different modality experts.

\subsection{Learning from Many Replica Neural Networks}
The proliferation of AI applications is reshaping the contours of the future knowledge graph of neural networks. Decentralized NNs is the study of knowledge transfer from different individual neural networks trained on separate local tasks to a global model. In a learning system comprising many replica neural networks with similar architecture and functions, the goal is to learn a global model that can generalize to unseen tasks without large-scale training \cite{survey}. In particular, we studied two practical problems in Decentralized NNs, i.e., learning with non-independent and identically distributed (non-iid) data and multi-domain data. 

Notably, non-iid refers to data samples across local models are not from the same distribution, which hinders the knowledge transfer between local models. To tackle the non-iid problem, we proposed the Segmented-Federated Learning (Segmented-FL) \cite{segmented} that employs periodic local model performance evaluation and learning group segmentation that brings neural networks training over similar data distributions together. Then, for each group, we train a different global model by transferring knowledge from the local models in the group. The global model can only passively observe the local model performance without access to the local data. We showed that the proposed method achieved better performance in tackling non-iid data of intrusion detection tasks compared to the traditional federated learning \cite{fedavg}.

On the other hand, multi-domain refers to data samples across local models are from different domains with domain-specific features. For example, an autonomous vehicle that learns to drive in a new city might leverage the driving data of other cities learned by different vehicles. Since different cities have different street views and weather conditions, it would be difficult to directly learn a new model based on the knowledge of the models trained on multi-domain data. This problem is closely related to multi-source domain adaptation, which studies the distribution shift in features inherent to specific domains that bring in negative transfer degrading a model's generality to unseen tasks. To this end, we proposed a new domain adaptation method that reduces feature discrepancy between local models and improves the global model’s generality to unseen tasks \cite{sun2022fedka}. We devised two components of embedding matching and global feature disentangler to align learned features of different local models such that the global model can learn better-refined domain-invariant features. Moreover, we found that a simple voting strategy that produces multiple predictions and generates pseudo-labels based on the consensus of local models could further improve the global model performance. The results of both image classification tasks and a natural language sentiment classification task showed that the proposed domain adaptation method could greatly improve the transfer learning of local models. 

\subsection{Building the Hierarchy of Neural Networks}
Hierarchical neural networks consist of multiple neural networks concreted in a form of an acyclic graph. An early theory of the global workspace theory (GWT) \cite{Baars1988} refers to multiple neural network models cooperating and competing in solving problems via a shared feature space for common knowledge sharing. Built upon the GWT, the conscious prior theory \cite{bengio2019} demonstrated the sparse factor graphs in space of high-level semantic variables and simple mapping between high-level semantic variables. 
%Several works like \cite{babyai} and \cite{goyal} are pushing the boundary in this direction. We consider a hierarchy of neural networks comprising two learning frameworks, i.e, fast learning and slow learning \cite{system12}. The fast learning framework comprises different individual modules while the slow learning framework is more like an attention mechanism for long-term planning. 
To study the hierarchy of neural networks, we proposed homogeneous learning for self-attention decentralized deep learning \cite{homo}. In particular, we devised a self-attention mechanism where a local model is selected as the meta for each training round and leverages reinforcement learning to recursively update a globally shared learning policy. The meta observes the states of local models and its surrounding environment, computing the expected rewards for taking different actions based on the observation. As mentioned in \cite{nature}, with a model of external reality and an agent's possible actions, it can try out various alternatives and conclude which is the best action using the knowledge of past events. The goal is to learn an optimized learning policy such that the Decentralized NNs systems can quickly solve a problem by planning and leveraging different local models' knowledge more efficiently. The results showed that the learning of a learning policy greatly reduced the total training time for an image classification task by 50.8\%. 

\subsection{Leveraging Different Modality Experts}

Information in the real world usually comes in different modalities. The degeneracy \cite{baby} in neural structure refers to any single function can be carried out by more than one configuration of neural signals and different neural clusters participate in several different functions. Intelligence systems build models of the world with different modalities where spatial concepts are generated via modality models. We demonstrate cross-modal learning in multimodal models \cite{clip}. Notably, we studied the Visual Question Answering (VQA) problem based on self-supervised learning \cite{unicon}. By leveraging the contrastive learning of different model components, we aimed to align the modality representations encouraging the similarity of the relevant component outputs while discouraging the irrelevant outputs. Such that the learning framework learns better-refined cross-modal representations for unseen VQA tasks based on the knowledge learned from different VQA tasks of local models.

\section{Anticipated Progress}

The vast majority of current neural networks lack sophisticated logical reasoning and action-planning modules. We aim to study a neuro-symbolic approach to improving the explainability and robustness of knowledge sharing in the Global Workspace (GW) of Decentralized NNs. Furthermore, we consider there are several necessary components for building such a neuro-symbolic learning framework, i.e., causal models and probabilistic Bayesian neural networks \cite{causal}, and associative memory like Hopfield Network \cite{hopfield}. In particular, we aim to tackle the tasks of visual grounding such as visual question answering and image captioning. In this regard, we will revisit and reintegrate the classical symbolic methods into the decentralized neural networks theory to improve the hierarchical reasoning of the meta agent for leveraging different modality expert models. The anticipated contribution is establishing a new learning framework to perform efficient causal discovery and inferences based on decentralized neural networks for improving generality in visual language modeling.

\bibliography{references}

\end{document}